\documentclass[runningheads]{llncs}
\usepackage{graphicx}

\usepackage{tikz}
\usepackage{comment}
\usepackage{amsmath,amssymb} 
\usepackage{color}
\usepackage[pagebackref=true,breaklinks=true,colorlinks,bookmarks=false]{hyperref}
\usepackage{url}            
\usepackage{booktabs}       
\usepackage{array}
\usepackage{algorithm, algorithmic}
\usepackage{graphicx}
\usepackage{amsfonts}
\usepackage{nicefrac}       
\usepackage{microtype}      
\usepackage{xcolor}         
\usepackage{multirow}
\usepackage{lipsum}
\usepackage{bm}
\usepackage{mathtools}
\usepackage{bbm}
\usepackage{xspace}
\usepackage{url}
\usepackage{cleveref}
\usepackage{soul}
\usepackage{ragged2e}
\usepackage{tabularx}
\usepackage[font={small}]{caption}
\usepackage{subcaption}
\usepackage{float}

\usepackage[accsupp]{axessibility}  

\newcommand\blfootnote[1]{%
    \begingroup
    \renewcommand\thefootnote{}\footnote{#1}%
    \addtocounter{footnote}{-1}%
    \endgroup
}

\newcommand{\etal}{\emph{et al}\ifx\@let@token.\else.\null\fi\xspace}

\begin{document}
\pagestyle{headings}
\mainmatter
\def\ECCVSubNumber{1541}  

\title{TL;DW? Summarizing Instructional Videos with Task Relevance \& Cross-Modal Saliency}

\titlerunning{Summarizing Instructional Videos}
%
\author{Medhini Narasimhan\inst{1,2}$^*$ \and
Arsha Nagrani\inst{2} \and
Chen Sun\inst{2,3} \and Michael Rubinstein\inst{2} \and Trevor Darrell\inst{1}$^\dagger$ \and Anna Rohrbach\inst{1}$^\dagger$ \and Cordelia Schmid\inst{2}$^\dagger$\\
}


\authorrunning{M. Narasimhan et al.}
%
\institute{$^{1}$UC Berkeley \qquad $^{2}$Google Research \qquad $^{2}$Brown University\\
\url{https://medhini.github.io/ivsum}}

\maketitle

\begin{center}
    \centering
    \includegraphics[width=0.8\textwidth]{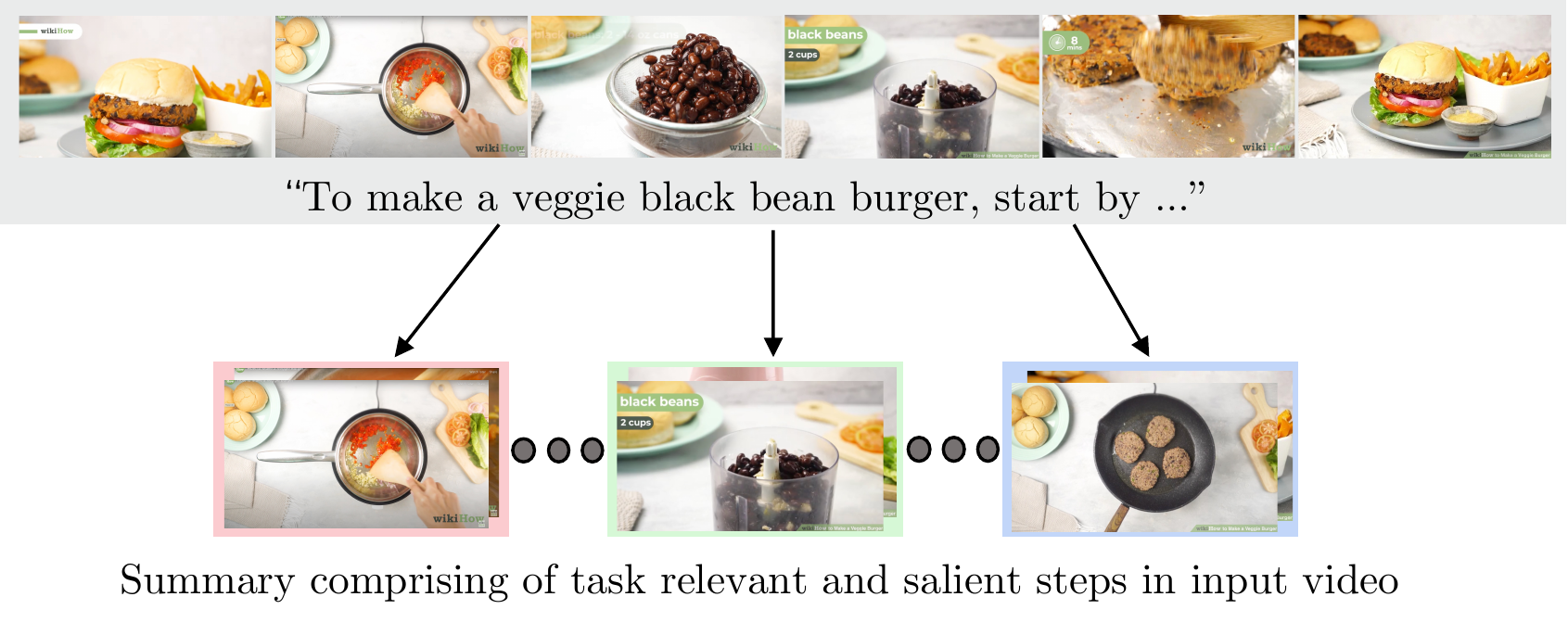}
    \captionof{figure}{\small{\textbf{Summarizing Instructional Videos} We introduce an approach for creating short visual summaries comprising steps that are most relevant to the task, as well as salient in the video, i.e.\ referenced in the speech. For example, given a long video on \emph{``How to make a veggie burger''} shown above, the summary comprises key steps such as \emph{fry ingredients, blend beans}, and \emph{fry patty}.}}
    \label{fig:teaser}
\end{center}

\begin{abstract}
YouTube users looking for instructions for a specific task may spend a long time browsing content trying to find the right video that matches their needs.\blfootnote{TL;DW? - Too Long; Didn't Watch?} Creating a visual summary (abridged version of a video) provides viewers with a quick overview and massively reduces search time.\blfootnote{$^*$Work done while an intern at Google Research. Correspondence to \texttt{medhini@berkeley.edu}} In this work, we focus on summarizing \emph{instructional} videos, an under-explored area of video summarization.\blfootnote{$^\dagger$Equal contribution.} In comparison to generic videos, instructional videos can be parsed into semantically meaningful segments that correspond to important steps of the demonstrated task. Existing video summarization datasets rely on manual frame-level annotations, making them subjective and limited in size. To overcome this, we first automatically generate \emph{pseudo summaries} for a corpus of instructional videos by exploiting two key assumptions: (i) relevant steps are likely to appear in multiple videos of the same task (\emph{Task Relevance}), and (ii) they are more likely to be described by the demonstrator verbally (\emph{Cross-Modal Saliency}). We propose an instructional video summarization network that combines a context-aware temporal video encoder and a segment scoring transformer. Using pseudo summaries as weak supervision, our network constructs a visual summary for an instructional video given only video and transcribed speech. To evaluate our model, we collect a high-quality test set, \emph{WikiHow Summaries}, by scraping WikiHow articles that contain video demonstrations and visual depictions of steps allowing us to obtain the ground-truth summaries. We outperform several baselines and a state-of-the-art video summarization model on this new benchmark.
\end{abstract}

\section{Introduction}


The search query \emph{``How to make a veggie burger?''} on YouTube yields thousands of videos, each showing a slightly different technique for the same task. It is often time-consuming for a first-time burger maker to sift through this plethora of video content. Imagine instead, if they could watch a compact visual summary of each video which encapsulates all semantically meaningful steps relevant to the task. Such a summary could provide a quick overview of what the longer video has to offer, and may even answer some questions about the task without the viewer having to watch the whole video. In this work, we propose a method to create such succinct visual summaries from long instructional videos.


Since our goal is to summarize videos, we consider prior work on generic~\cite{Gygli14,Song15} and query-focused \cite{sharghi2017query} video summarization. Generic video summarization datasets \cite{Gygli14,Song15} tend to contain videos from \textit{unrestricted domains} such as sports, news and day-to-day events. Given that annotations are obtained manually, the notion of what constitutes a good summary is subjective, and might differ from one annotator to the next. Query-focused video summarization partially overcomes this subjectivity by allowing users to customize a summary by specifying a natural language query~\cite{sharghi2017query,narasimhan2021clip}. However, both generic and query-focused approaches require datasets to be annotated manually at a per-frame level. This is very expensive, resulting in very small-scale datasets (25-50 videos) with limited utility and generalization. 


Here, we focus on a specific domain -- that of instructional videos \cite{tang2019coin,zhukov2019cross,miech2019howto100m}. We argue that a unique characteristic of these videos is that a summary can be clearly defined as a minimally sufficient \textit{procedural} one, i.e., it must include the steps necessary to complete the task (see Fig.~\ref{fig:teaser}). To circumvent having to manually annotate our training data, we use an unsupervised algorithm to obtain weak supervision in the form of pseudo ground-truth summaries for a large corpus of instructional videos. We design our unsupervised objectives based on two hypotheses: (i) steps that are relevant to the task will appear across multiple videos of the same task, and (ii) salient steps are more likely to be described by the demonstrator verbally. In practice, we segment the video and group individual segments into steps based on their visual similarity. Then we compare the steps across videos of the same task to obtain \emph{task relevance scores}. We also transcribe the videos using Automatic Speech Recognition (ASR) and compare the video segments to the transcript. We aggregate these \emph{task relevance} and \emph{cross-modal scores} to obtain the \emph{importance scores} for all segments, i.e., our pseudo ground-truth summary. 

\looseness=-1
Next, given an input video and transcribed speech, we train an instructional video summarization network (\emph{IV-Sum}). IV-Sum learns to assign scores to short \textit{video segments} using 3D video features which capture temporal context. Our network consists of a video encoder that learns context-aware temporal representations for each segment and a segment scoring transformer (SST) that then assigns importance scores to each segment. Our model is trained end-to-end using the importance scores from the pseudo summaries. Finally, we concatenate the highest scoring segments to form the final video summary. 

\looseness=-1
While we can rely on pseudo ground-truth for training, we collect a clean, manually verified test set to evaluate our method. Since manually creating a labeled test set from scratch would be extremely expensive, we find a solution in the form of the WikiHow resource\footnote{\url{https://www.wikihow.com/}}. WikiHow articles often contain a link to an instructional video and a set of human-annotated steps present in the task along with corresponding images or short clips. To construct our test set (referred to as \emph{WikiHow Summaries}), we automatically localize these images/clips in the video. We obtain localized segments for the images (using a window around the localized frame) and clips, and stitch the segments together to create a summary.  This provides us with binary labels for each frame which serve as ground-truth annotations. We evaluate our model on \emph{WikiHow Summaries} and compare it to several baselines and the state-of-the-art video summarization model CLIP-It~\cite{narasimhan2021clip}. Our model surpasses prior work and several baselines on three standard metrics (F-Score, Kendall~\cite{Kendall}, and Spearman~\cite{Spearman} coefficients).

\looseness=-1
To summarize (pun intended), we introduce an approach for summarizing instructional videos that involves training our \emph{IV-Sum} model on pseudo summaries created from a large corpus of instructional videos. \emph{IV-Sum} learns to rank different segments in the video by learning context-aware temporal representations for each segment and a segment scoring transformer that assigns scores to segments based on their task relevance and cross-modal saliency. Our method is weakly-supervised (it only requires the task labels for videos), multimodal -- uses both video and speech transcripts, and is scalable to large online corpora of instructional videos. We collect a high-quality test set, \emph{WikiHow Summaries} for benchmarking instructional video summarization, which will be publicly released. Our model outperforms state-of-the-art video summarization methods on all metrics. Compared to the baselines, our method is especially good at capturing task relevant steps and assigning higher scores to salient frames, as seen through qualitative analysis.

\section{Related Work}

We review several lines of work related to summarization of instructional videos.

\noindent\textbf{Generic Video Summarization.} This task involves creating abridged versions of generic videos by stitching together short important clips from the original video~\cite{he16,Mahasseni17,narasimhan2021clip,park2020sumgraph,Rochan19,Yuan2019,Zhang16,Zhang18,zhao2018hsa}. Some of the more recent methods attempt to learn contextual representations to perform video summarization, via attention mechanism~\cite{accvw18}, graph based~\cite{park2020sumgraph} or transformer-based~\cite{narasimhan2021clip} methods. Representative datasets include SumMe~\cite{Gygli14} and TVSum~\cite{Song15}, where the ground-truth summaries were created by annotators assigning scores to each frame in the video, which is highly time consuming and expensive. As a consequence, the generic video summarization datasets are small and the quality of the summaries is often very subjective. Here, we focus on instructional videos which contain structure in the form of task steps, thus we have a clear definition of what a good summary should contain - a set of necessary steps for performing that specific task.

\noindent\textbf{Query Focused Video Summarization.} To address the subjectivity issues with Generic Summarization, Query Focused Video Summarization allowed for having user defined natural language queries to customize the summaries~\cite{kanehira2018aware,sharghi2017query,wei2018video}. A representative dataset is Query Focused Video Summarization~\cite{sharghi2016query}; it is very small and the queries correspond to a very narrow set of objects. In contrast, our task is large and we do not rely on any additional user input. 

\noindent\textbf{Step Localization.} Step localization (also known as temporal action segmentation) is a related albeit distinct task. It typically implies predicting temporal boundaries of steps when the step labels~\cite{richard2018action,tang2019coin,zhukov2019cross} and even their ordering~\cite{bojanowski2014weakly,chang2019d3tw,ding2018weakly,huang2016connectionist,kuehne17weakly,richard2017weakly} are given. Representative datasets, COIN~\cite{tang2019coin} and CrossTask~\cite{zhukov2019cross} consist of instructional videos and a fixed set of steps for each task (from the WikiHow resource), and the task is to localize these steps in the video. Our task is different in that we are only given a video without corresponding input steps. Our model learns to pick out segments that correspond to relevant and salient steps in order to construct a video summary. We discuss and illustrate the shortcomings of the step localization annotations in Sec.~\ref{sec:exp} and Fig.~\ref{fig:oursvsstepL}.

\noindent\textbf{Unsupervised Parsing of Instructional Videos.} Closest to ours is the line of work on unsupervised video parsing and segmentation that discovers steps in instructional videos in an unsupervised manner~\cite{alayrac2016unsupervised,fried2020learning,kukleva2019unsupervised,sener2015unsupervised,sener2018unsupervised}. However, these works - (1) do not focus on video summarization, thus they might miss some salient steps in video, (2) often use very small datasets for training and evaluation that do not capture the broad range of instructional videos found in, e.g., COIN~\cite{tang2019coin} and CrossTask~\cite{zhukov2019cross}.

\section{Summarizing Instructional Videos}

\begin{figure*}[t]
    \centering
    \includegraphics[width=\textwidth]{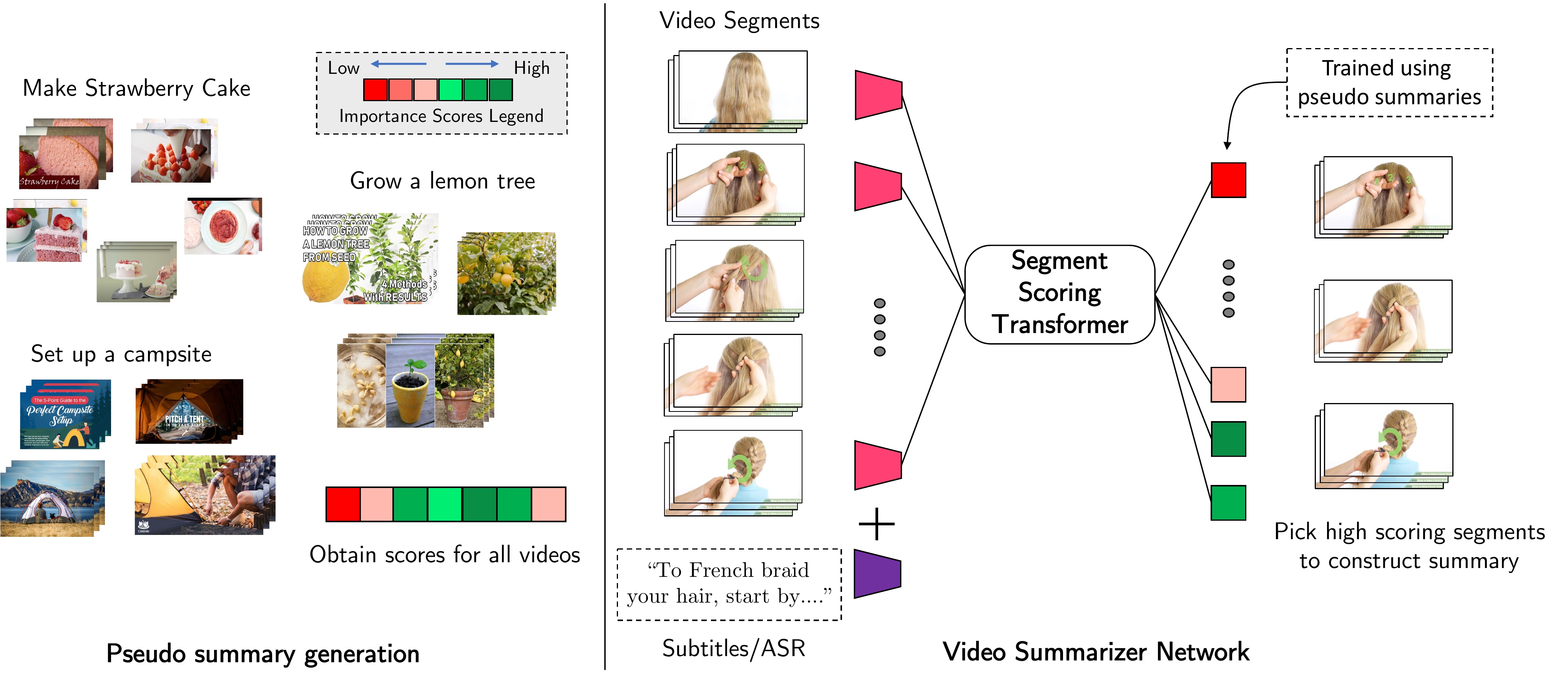}
    \caption{\textbf{Summarizing Instructional Videos.} We first obtain pseudo summaries for a large collection of videos using our weakly supervised algorithm (more details in Fig.~\ref{fig:pseudodgt}). Next, using the pseudo summaries as weak-supervision, we train our Instructional Video Summarizer (\emph{IV-Sum}). It takes an input video along with the corresponding ASR transcript and learns to assign importance scores to each segment in the video. The final summary is a compilation of the high scoring video segments.
    }
    \label{fig:approach}
\end{figure*}

\textbf{Overview.} We propose a novel approach for constructing visual summaries of instructional videos. An instructional video typically consists of a visual demonstration of a specific task, e.g. \emph{``How to make a pancake?''}. Our goal is to construct a visual summary of the input video containing only the steps that are crucial to the task and salient in the video, i.e. referenced in the speech. Fig.~\ref{fig:approach} illustrates an outline of our approach. Our instructional video summarization pipeline consists of two stages - (i) first, we use a weakly supervised algorithm to generate pseudo summaries and frame-wise importance scores for a large corpus of instructional videos, relying only on the task label for each video (ii)~next, using the pseudo summaries as supervision, we train an instructional video summarization network which takes as input the video and the corresponding transcribed speech and learns to assign scores to different segments in the input video. The network consists of a video encoder and a segment scoring transformer (SST) and is trained using the importance scores of the pseudo summaries. The final summary is constructed by selecting and concatenating the segments with high importance scores. We first describe our pseudo summary generation algorithm, followed by details on our instructional video summarizer (\emph{IV-Sum}), and the inference procedure. 

\subsection{Generating Pseudo Summaries}
\label{sec:pseudogt}
Since manually collecting annotations for summarization is expensive and time consuming, we propose an automatic weakly supervised approach for generating summaries that may contain noise but have enough valuable signal for training a summarization network. The main intuition behind our pseudo summary generation pipeline is that given many videos of a task, steps that are crucial to the task are likely to appear across multiple videos (task relevance). Additionally, if a step is important, it is typical for the demonstrator to speak about this step either before, during, or after performing it. Therefore, the subtitles for the video obtained using Automatic Speech Recognition (ASR) will likely reference these key steps (cross-modal saliency). These two hypotheses shape our objectives for generating pseudo summaries.

\begin{figure*}[t]
    \centering
    \includegraphics[width=\textwidth]{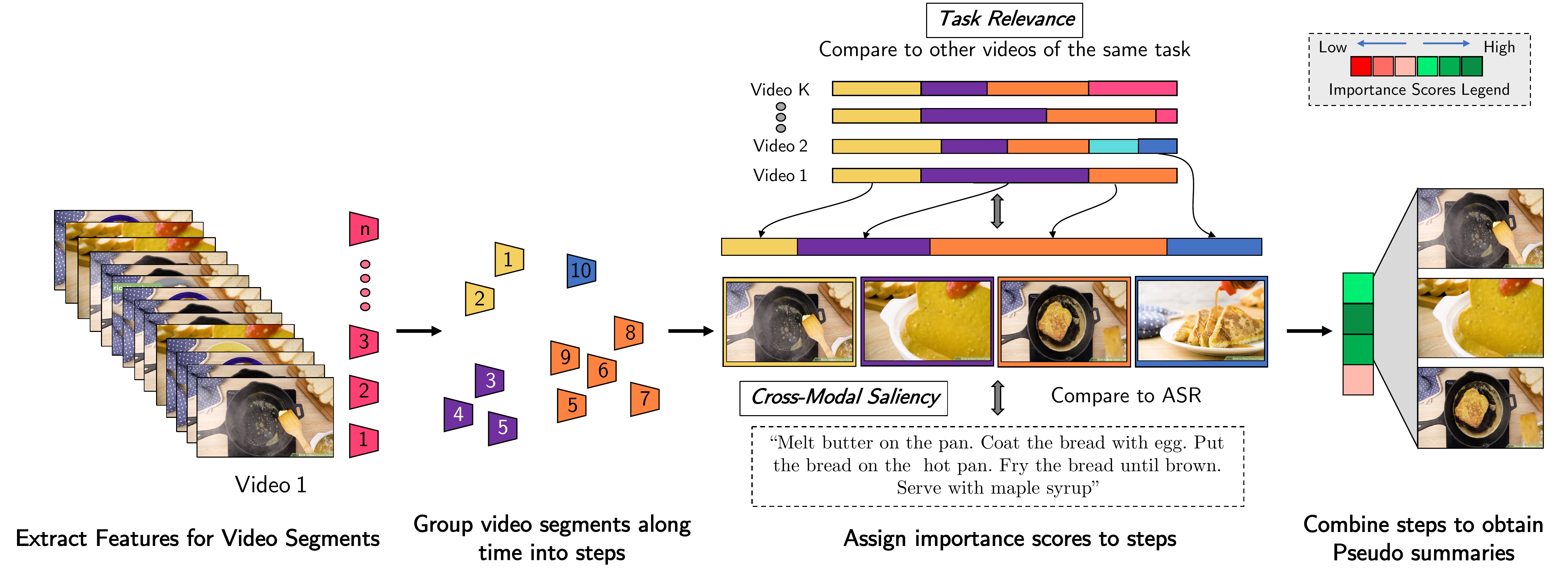}
    \caption{\textbf{Pseudo Summary Generation.} To generate the pseudo summary, we first uniformly partition the video into segments, then group the segments based on visual similarity into steps (shown in different colors), assign \emph{importance scores} to steps based on \emph{Task Relevance} and \emph{Cross-Modal Saliency}, and then pick high scoring steps to obtain pseudo summaries. }
    \label{fig:pseudodgt}
\end{figure*}

\noindent\textbf{Task Relevance.} We first group videos based on the task. Say videos $V_i, i \in [1, \dots \mathcal{K}]$ are $\mathcal{K}$ videos from the same task, as shown in Fig.~\ref{fig:pseudodgt}. For a given video, we divide it into $\mathcal{N}$ equally sized non-overlapping segments $s_i, i \in [1, \dots \mathcal{N}]$ and embed each segment using a pre-trained 3D CNN video encoder $g_{vid}$~\cite{miech2020end}. We merge segments along the time axis based on their dot-product similarity, i.e. if similarity of a segment to the one prior to it is greater than a threshold, the two are grouped together and the joint feature representation is an average of the feature representation of the two segments. The threshold for similarity is heuristically set to be 90\% of the maximum similarity between any two segments in the video. We call these merged segments \emph{steps}, as they typically correspond to semantic steps as we show through qualitative results in supplemental. We do this for all $\mathcal{K}$ videos in the task, and then compare each step to all the $\mathcal{S}$ steps across all $\mathcal{K}$ videos of the task. We assign \emph{task relevance scores} $\text{trs}_{S_i}$, to each step $S_i, i \in \mathcal{S}$ based on its visual similarity to all the $\mathcal{S}$ steps from all $K$ videos of this task, as shown below:
$$
\text{trs}_{S_i} = \frac{1}{|\mathcal{S}|}\sum_{j \in \mathcal{S}}{g_{vid}(S_i) \cdot g_{vid}(S_j)} 
$$

\looseness=-1
\noindent\textbf{Cross-Modal Saliency.} We also compare each video step to each sentence in the transcript of the same video. This enforces our idea that if a step is important, it will likely be referenced in the speech. To do this, we encode both, the input segments and the transcript sentences, using a pre-trained video-text model where the video and text streams are trained jointly using MIL-NCE loss~\cite{miech2020end}. Each visual step is assigned a \emph{cross-modal score} by averaging its similarity over all the sentences. 

Each step (and all the segments in it) is then assigned an importance score that is an average of the \emph{task relevance} and the \emph{cross-modal scores}. This constitutes our pseudo summary scores. For any given video, the top $t\%$ highest scoring steps are retained to be a part of the summary.  

\subsection{Instructional Video Summarizer (\emph{IV-Sum})}
\label{sec:ivsum}
Recall that our goal is to construct a visual summary of any instructional video by picking out the important steps in it, without having to rely on other videos of the same task or the task label. To do this, we use the pseudo summaries generated above as weak supervision to train \emph{IV-Sum}, which learns to assign importance scores to individual segments in the video using only the information in the video and the corresponding transcripts as seen in Fig.~\ref{fig:approach}. While some prior summarization methods operate on independent frames~\cite{narasimhan2021clip,park2020sumgraph}, \emph{IV-Sum} operates on non-overlapping segments $s_i, i \in [1, \dots \mathcal{N}]$, and learns \emph{context-aware temporal representations} using a 3D CNN video encoder $f_{\text{vid}}$. The transcript is projected onto the same embedding space using a text encoder $f_{\text{text}}$, and the text representations are concatenated individually to each of the segments. To contextualize information across several segments, we use a segment scoring encoder-only transformer~\cite{vaswani2017attention} $f_{\text{trans}}$ with positional embeddings, that assigns importance scores  $Y'_{s_i}$ to each segment as shown in Eq.~\ref{eq:ivsum}. The network is trained using supervision from the importance scores of the pseudo summaries $Y_{s_i}$, using Mean-Squared Error Loss as shown in Eq.~\ref{eq:mse}.  

\begin{align}
    \label{eq:ivsum}
    Y'_{s_i} &= f_{\text{trans}}(\text{concat} \: \left(f_{\text{text}}(\text{transcript}),f_{\text{vid}}(s_i))\right) \; \forall \; i \in \mathcal{N} \\
    \mathcal{L}_{\text{IV-Sum}} &= \sum_{i \in \mathcal{N}} \text{MSE}\: (Y'_{s_i}, Y_{s_i})
    \label{eq:mse}
\end{align}

During inference, we sort the segments based on the predicted scores and assign the label 1 to the top $t\%$ of the segments, and the label 0 to the remaining ones. When a segment is assigned a label, all the frames in the segment also get assigned the same label. The summary is constructed by stitching together all the frames with label 1.   
   
\section{Instructional Video Summarization Datasets}
\label{sec:WikiHow}
We describe the details of the data collection process for the annotations used in our work --- \emph{Pseudo Summaries} annotations for training and the \emph{WikiHow Summaries} annotations for evaluation.   

\noindent\textbf{Pseudo Summaries Training Dataset.} As described in Sec.~\ref{sec:pseudogt}, we use the pseudo summary generation process for creating our training set. We use the videos and task annotations from COIN~\cite{tang2019coin} and CrossTask~\cite{zhukov2019cross} datasets for creating our training datasets. \\
\noindent\textbf{COIN:} COIN consists of 11K videos related to 180 tasks. As this is a dynamic YouTube dataset, we were able to obtain 8,521 videos at the time of this work. \\
\noindent\textbf{Cross-Task:} CrossTask consists of 4,700 instructional videos (of which we were able to access 3,675 videos) across 83 different tasks. \\
\noindent\textbf{Pseudo Summaries:} We combined the two datasets to create pseudo summaries comprising of 12,160 videos, whilst using the videos that were common to both datasets only once. They span 263 different tasks, have an average length of 3.09 minutes, and in total comprise of 628.53 hours of content. The summary videos that were constructed using our pseudo ground-truth generation pipeline are 1.71 minutes long on an average, with each summary being 60\% of the original video. While it is possible to construct pseudo summaries using the step-localization annotations, we show in Sec.~\ref{sec:exp} that such summaries may miss important steps or do not pick up on steps that are salient in the video. Moreover, our pseudo summary generation mechanism is weakly-supervised, requiring only task annotations and no step-localization annotations.

\begin{table*}[t]
\centering
\caption{\textbf{Instructional Video Summarization Datasets Statistics.} $\dagger$ Our \emph{WikiHow Summaries} dataset was created automatically using a scalable pipeline, but manually verified for correctness.}
\begin{tabular}{@{}lccccc@{}}\toprule
    & TVSum & SumMe & Pseudo Summaries & WikiHow Summaries \\\midrule 
    Number of videos & 50 & 25 & 12160 & 2106 \\
    Annotation & Manual & Manual & Automatic & Manually verified$\dagger$\\
    Number of Tasks/Categories & 10 & 25 & 185 & 20 \\
    Total Input Duration (Hours) & 3.5 & 1.0 & 628.53 & 42.94 \\  
    \bottomrule
    \end{tabular}
\label{tab:stat}
\end{table*}

\noindent\textbf{WikiHow Summaries Dataset.} To provide a test bed for instructional video summarization, we automatically create and manually verify \emph{WikiHow Summaries}, a video summarization dataset consisting of 2,106 input videos and summaries, where each video describes a unique task. Each article on the \href{https://www.wikihow.com/Video}{WikiHow Videos} website consists of a main instructional video demonstrating a task that often includes promotional content, clips of the instructor speaking to the camera with no visual information of the task, and steps that are not crucial for performing the task. Viewers who want an overview of the task would prefer a shorter video without all of the aforementioned irrelevant information. The WikiHow articles (e.g., see \href{https://www.wikihow.com/Make-Sushi-Rice}{How to Make Sushi Rice}) contain exactly this: corresponding text that contains all the important steps in the video listed with accompanying images/clips illustrating the various steps in the task. These manually annotated articles are a good source for automatically creating ground-truth summaries for the main videos. We obtain the summaries and the corresponding labels and importance scores using the following process (see supp. for an overview figure):

\noindent\textbf{1. Scraping WikiHow videos.} We scrape the \href{https://www.wikihow.com/Video}{WikiHow Videos} website for all the long instructional videos along with each step and the images/video clips (GIFs) associated with the step. 

\noindent\textbf{2. Localizing images/clips.} We automatically localize these images/clips in the main video by finding the closest match in the video. To localize an image, we compare ResNet50~\cite{he2016deep} features of the image and to that of all the frames in the video. The most similar frame is selected and this step is localized in the input video to a 5 second window centered around the frame. If the step contains a video clip/GIF, we localize the first frame of the video clip/GIF in the input video by similarly comparing ResNet features, as above, and the localization is set to be the length of the step video clip.   

\noindent\textbf{3. Ground-truth summary from localized clips.} We stitch the shorter localized clips together to create the ground truth summary video. Consequently, we assign labels to each frame in the input video, depending on whether it belongs to the input summary (label 1) or not (label 0). To obtain importance scores, we partition each input video into equally sized segments (same as in Sec.~\ref{sec:ivsum}) and compute the importance score for each segment to be the average of the labels assigned to the individual frames in the segment. 

\noindent\textbf{4. Manual verification.} We verified that the summaries are at least 30\% of the original video and manually fixed summaries that were extremely short/long.

\noindent\emph{Online Longevity and Scalability.} We note that a common problem plaguing YouTube datasets today is shrinkage of datasets as user uploaded videos are taken down by users (eg. Kinetics~\cite{i3d}). WikiHow articles are less likely to be taken down, and this is an actively growing resource as new How-To videos are released and added (25\% growth since we collected the data). Hence there is a potential to continually increase the size of the dataset.

For each video, we provide the following: (i) frame-level binary labels (ii) the summary formed by combining the frames with label 1 (iii) segment-level importance scores between 0 and 1, which are computed as an average of the importance scores for all the frames in the segment (iv) the localization of the visual steps in the video (i.e. the frames associated with each step). We also scrape natural language descriptions of each step as a bonus that could be useful for future work. We divide our WikiHow dataset into 768 validation and 1,339 test videos. Tab.~\ref{tab:stat} shows the statistics of both our datasets. Both datasets are much larger in size compared to existing generic video summarization datasets, contain a broader range of tasks, and are scalable.

\section{Experiments}
\label{sec:exp}
Next, we describe the experimental setup and evaluation for instructional video summarization. We compare our method to several baselines, including CLIP-It~\cite{narasimhan2021clip}, the state-of-the-art on generic and query-focused video summarization.

\noindent\textbf{Implementation Details.} For the video and text encoders, we use an S3D~\cite{xie2018rethinking} network, initialized with weights from pre-training on HowTo100M~\cite{miech2019howto100m} using the MIL-NCE loss~\cite{miech2020end}. We fine-tune the \emph{mixed\_5*} layers and freeze the rest. The segment scoring transformer  is an encoder consisting of 24 layers and 8 heads and is initialized randomly. The network is trained using the Adam optimizer~\cite{kingma2014adam}, with learning rate of 0.01, and a batch size of 24. We use Distributed Data Parallel to train for 300 epochs across 8 NVIDIA RTX 2080 GPUs. Additional implementation details are mentioned in supplemental. 

\noindent\textbf{Metrics.} To evaluate instructional video summaries, we follow the evaluation protocol used in past video summarization works~\cite{Zhang16CVPR,park2020sumgraph,narasimhan2021clip} and report Precision, Recall and F-Score values. 
As described in Sec.~\ref{sec:WikiHow}, each video in the \emph{WikiHow Summaries} dataset contains the ground-truth labels $Y_l$ (binary labels for each frame in the video) and the ground-truth scores $Y_s$ (importance scores in the range [0-1] for each segment in the video). 
%
We compare the binary labels predicted for the frames in the video $Y'_l$, to the ground truth labels $Y_l$, and measure F-Score, Precision and Recall, as defined in prior summarization works~\cite{Rochan18,Rochan19}. 



While these scores assess the quality of the predicted frame-wise binary labels, to assess the quality of the predicted segment-wise importance scores $Y'_s$, we follow Otani \etal~\cite{Otani19}, and report results on the rank-based metrics Kendall's $\tau$~\cite{Kendall} and Spearman's $\rho$~\cite{Spearman} correlation coefficients. 
We first rank the video frames according to the generated importance scores $Y'_s$ and the ground-truth importance scores $Y_s$. We then compare the generated ranking to each ground-truth ranking of video segments for each video obtained from the frame-wise binary labels as described in Sec.~\ref{sec:WikiHow}. The final correlation score is computed by averaging over the individual scores for each video. 

\begin{table*}[t]
    \centering
    \small
    \caption{\small{\textbf{Instructional Video Summarization results on \emph{WikiHow Summaries.}} We compare F-Score, Kendall and Spearman correlation metrics of our method IV-Sum, to all the baselines. Our method achieves state-of-the-art on all three metrics.}}
    \begin{tabularx}{\textwidth}{lXXXXXXX}\toprule
    \multicolumn{4}{c}{\multirow{2}{*}{Method}} & \multicolumn{2}{c}{F-Score} & $\tau$~\cite{Kendall} & $\rho$~\cite{Spearman} \\ 
    \cmidrule(l{-1pt}r{10pt}){5-6} \multicolumn{4}{c}{} & Val & Test & Test & Test\\ \midrule
    & ASR & RGB & Pseudo & & \\
    \midrule
        Frame Cross-Modal Similarity & \checkmark & \checkmark & - & 52.8 & 53.1 & 0.022 & 0.051 \\
        Segment Cross-Modal Similarity & \checkmark & \checkmark & - & 55.1 & 55.5 & 0.034 & 0.060 \\
        Step Cross-Modal Similarity & \checkmark & \checkmark & - & 57.9 & 58.3 & 0.037 & 0.061 \\
        CLIP-It with captions~\cite{narasimhan2021clip} & - & \checkmark & - & 22.5 & 22.1 & 0.036 & 0.064\\
        CLIP-It with ASR~\cite{narasimhan2021clip} & \checkmark & \checkmark & - & 27.9 & 27.2 & 0.055 & 0.088\\ \midrule
        CLIP-It with ASR & \checkmark & \checkmark & \checkmark & 62.5 & 61.8 & 0.093 & 0.191 \\ 
        IV-Sum without ASR & - & \checkmark & \checkmark & 65.8 & 65.2 & 0.095 & 0.202\\ 
        \textbf{IV-Sum} & \checkmark & \checkmark & \checkmark & \textbf{67.9} & \textbf{67.3} & \textbf{0.101} & \textbf{0.212}\\ 
        \bottomrule
    \end{tabularx}  
\label{tab:1}
\end{table*}

\noindent\textbf{Baselines.} We compare our method to the state-of-the-art video summarization model CLIP-It~\cite{narasimhan2021clip}. To validate the need for pseudo summaries, we construct three unsupervised baselines as alternatives to our pseudo summary generation algorithm. We first describe the three unsupervised baselines. 

\noindent \textbf{Frame Cross-Modal Similarity.} We sample frames (at the same FPS used by our method) from an input video and compute the similarity between CLIP~(ViT-B/32)~\cite{radford2021learning} frame embeddings and CLIP text embeddings of each sentence in the transcript. The embeddings do not encode temporal information but leverage the priors learned by the CLIP model. Based on the scores assigned to each frame, we threshold $t\%$ of the higher scoring frames to be part of the summary. Frame scores are propagated to the segments they belong to, and the summary is a compilation of the chosen segments.

\noindent\textbf{Segment Cross-Modal Similarity.} We uniformly divide the video into segments and compute MIL-NCE~\cite{miech2020end} video features for each segment. We embed each sentence in the transcript to the same feature space using the MIL-NCE text encoder. We compute the pairwise similarity between all video segments and the sentences, and average over sentences to obtain a score for each segment. Our intuition is that since demonstrators typically describe the important steps shortly before, after or while performing them, a high similarity between the visuals and transcripts would directly correlate with the significance of the step. We filter $t$\% of the highest scoring segments, where $t$ is determined heuristically using the \emph{WikiHow Summaries} validation set and is consistent across all baselines and our model. The filtered segments are stitched together to form the summary.

\noindent\textbf{Step Cross-Modal Similarity.} 
We first group segments into steps and then compare them  to the ASR transcripts. For this we employ the technique described in Sec.~\ref{sec:pseudogt}, i.e. we extract MIL-NCE features for the video segments and group them together based on their similarity to form steps.\footnote{Since we process a single input video (not multiple videos per task), we can not use the Task Relevance component.} The embedding for a step is set to be the average of all the segment embeddings in it. If a step is similar to the transcript, all the segments in that step are chosen to be part of the summary. This baseline is the closest to our pseudo summary generation algorithm. 

\begin{figure*}[t]
    \centering
    \includegraphics[width=\textwidth]{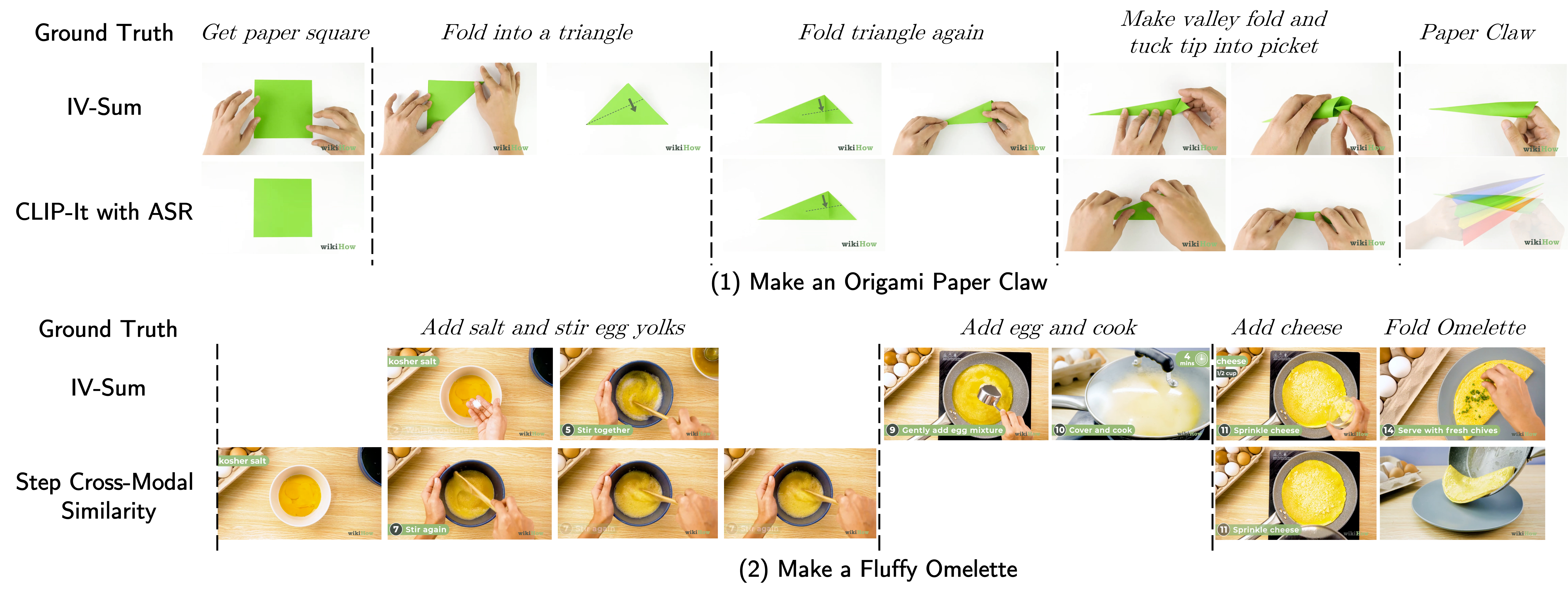}
    \caption{\textbf{Qualitative comparisons to baselines.} We show the steps in the ground-truth as text (note we never train with step descriptions, these are shown here simply for illustrative purposes) and compare frames selected in summaries generated by our method IV-Sum, CLIP-It with ASR, and Step Cross-Modal Similarity. In (1), CLIP-It misses steps which are deemed important by our method (\emph{``Fold into a triangle''}) and assigns higher scores to less salient frames for the step (\emph{``Make valley fold and tuck tip into picket''}) where neither the valley fold nor the picket are clearly visible. In (2), Step Cross-Modal Similarity misses (\emph{``Add egg and cook''}) and selects too many redundant frames for the step (\emph{``Add salt and stir egg yolks''}).}
    \label{fig:q1}
\end{figure*}

\begin{figure*}[t]
    \centering
    \includegraphics[width=0.8\textwidth]{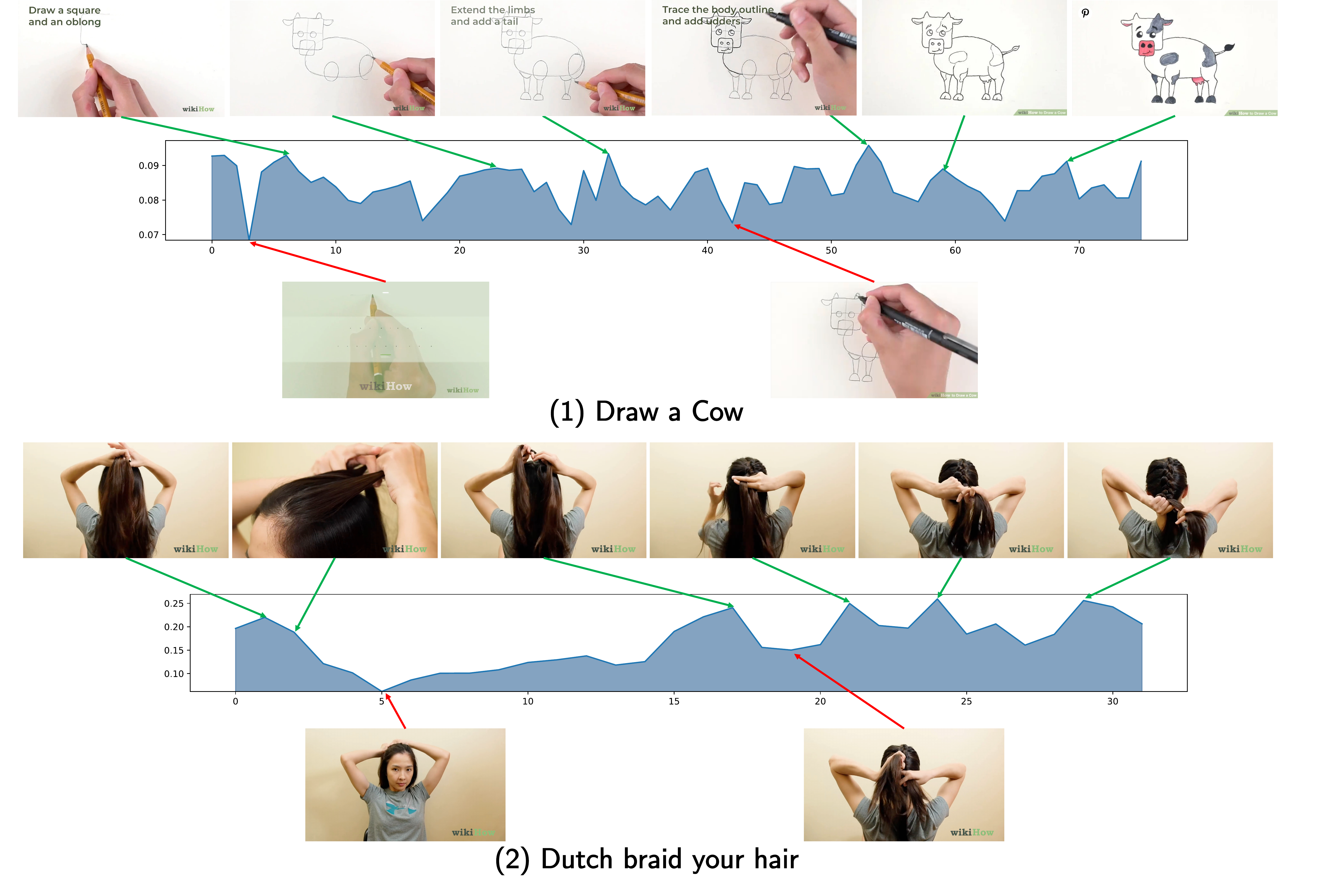}
    \caption{\textbf{Qualitative results.} We show summaries from our method IV-Sum along with the predicted importance scores. The \textcolor{green}{green} and \textcolor{red}{red} arrows point to frames that were assigned a high and low scores, respectively. Our model correctly assigns higher scores to frames from all the steps that are relevant and lower scores to frames which aren't crucial to the task (as in (1)) and frames which don't belong to a step (as in (2)).}
    \label{fig:q2}
\end{figure*}

Next, we describe the CLIP-It baseline and ablations, trained with supervision. 

\looseness=-1
\noindent\textbf{CLIP-It with captions.} We evaluate CLIP-It~\cite{narasimhan2021clip} trained on TVSum~\cite{Song15}, SumMe~\cite{Gygli14}, OVP~\cite{OVP}, and YouTube~\cite{de11} against our \emph{WikiHow Summaries}. We use the same protocol as in CLIP-It for evaluation and describe further details in supplemental. For language-conditioning, we follow CLIP-It and generate captions for the \emph{WikiHow Summaries} dataset using BMT~\cite{iashin2020better}; we feed these as input to the CLIP-It model. 

\noindent\textbf{CLIP-It with ASR transcripts.} We evaluate the same CLIP-It model above by replacing captions with ASR transcripts, so as to allow for a fair comparison with the baselines and our method, IV-Sum which use ASR transcripts. 

\noindent\textbf{CLIP-It with ASR transcripts trained on Pseudo Summaries.} We train CLIP-It from scratch on our Pseudo-GT Summaries dataset using ASR transcripts from the videos in place of captions. 

\noindent\textbf{Quantitative Results.} We compare the baselines to the two versions of IV-Sum, one with ASR transcripts and another without. To train IV-Sum without transcripts, we simply eliminate the text encoder ($f_{text}$) in Eq.~\ref{eq:ivsum} and pass only the visual embeddings of the individual segments to the transformer. We report F-Score, Kendall's $\tau$ and Spearman's $\rho$ coefficients in Tab.~\ref{tab:1}. As seen, IV-Sum (both with and without ASR transcripts), outperforms all the baselines on all metrics. Particularly, we achieve notable improvements on the correlation metrics that compare the saliency scores, attesting to our model's capabilities to assign higher scores to segments that are more relevant.  We also observe that CLIP-It trained using the pseudo summaries generated by our method has a strong boost in performance compared to CLIP-It trained on generic video summarization datasets, reinforcing the effectiveness of our pseudo summaries for training. The best method among the unsupervised ones is Step Cross-Modal Similarity, a ``reduced'' version of our pseudo summary generation method.

\noindent\textbf{Qualitative Results.} We present qualitative results in Fig.~\ref{fig:q1}. We show frames in the summaries generated by our method IV-Sum, CLIP-It with ASR transcripts (trained on generic video summarization datasets), and Step Cross-Modal Similarity. We also list the steps in the ground-truth as text (for illustrative purposes). In Fig.~\ref{fig:q1} (1), CLIP-It misses the step \emph{``Fold into a triangle''}, as it optimizes for diversity among the frames and was trained on a small dataset that does not generalize well to our domain. It also picks the less salient frames for the step \emph{``Make valley fold and tuck tip into picket''}, whereas our model correctly identifies all the steps and assigns higher scores to the more salient frames. The summary from the Step Cross-Modal Similarity baseline, shown in Fig.~\ref{fig:q1} (2), assigns high scores to several redundant frames (\emph{``Add salt and stir egg yolks''}), but misses ``Add egg and cook''.

\begin{figure*}[t]
    \centering
    \includegraphics[width=\textwidth]{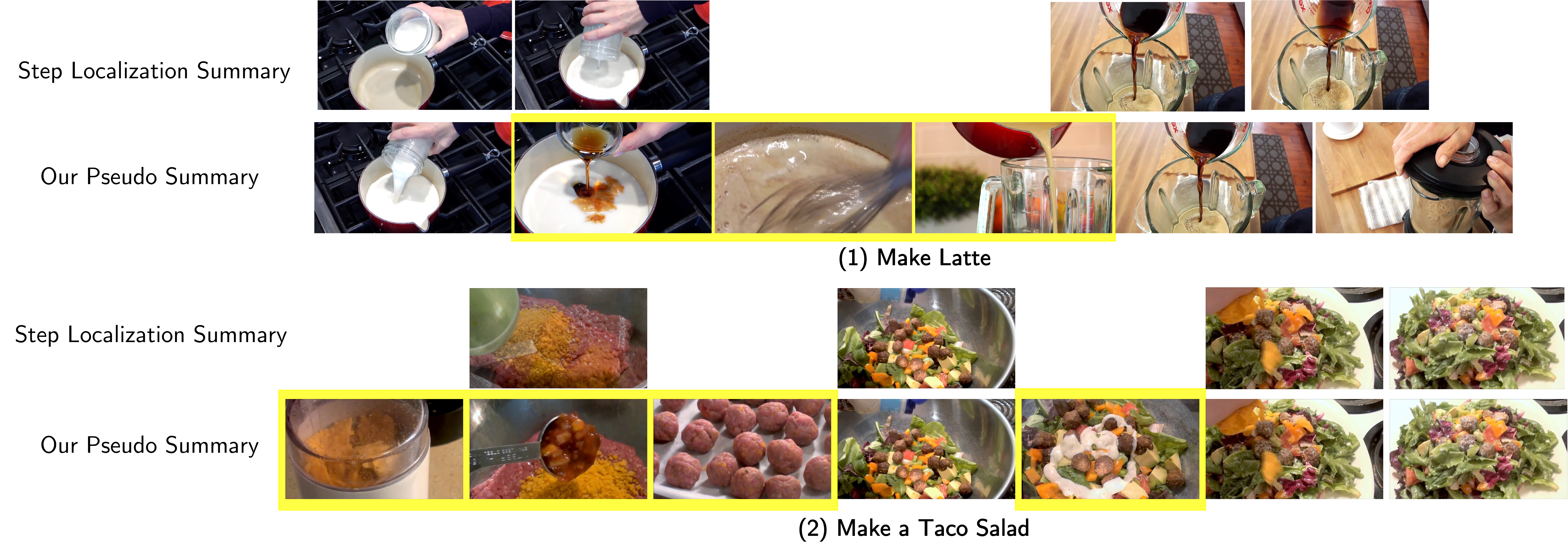}
    \caption{\textbf{Pseudo summaries vs step-localization annotations.} We compare frames in our automated pseudo summary to the step localization manual annotations, aligned temporally. Frames corresponding to steps that are identified by our method but missed by step localization are highlighted in \textcolor{yellow}{yellow}.}
    \label{fig:oursvsstepL}
\end{figure*}

Fig.~\ref{fig:q2} shows results from our method along with the predicted frame-wise importance scores. The \textcolor{green}{green} and \textcolor{red}{red} arrows point to frames that are assigned the highest and lowest scores by our method, respectively. As seen, our method assigns high scores to frames in task relevant and salient steps and low scores to frames which aren't crucial to the step, like in Fig.~\ref{fig:q2} (1), or do not belong to a step, like in Fig.~\ref{fig:q2} (2) where the person is talking to the camera.

\begin{table*}[t]
 \centering
\caption{\small{\textbf{Pseudo Summary Variations.} We report results on two variations of generating the pseudo summaries: (i) ablating the objectives (ii) using step localization annotations to generate pseudo summaries.}}  
\begin{subtable}{0.48\textwidth}{
\centering
\caption{\small{\textbf{Ablating objectives.} We ablate the two objectives in our pseudo summary generation pipeline.}}
\begin{tabular}{lcc}\toprule
    \textbf{Method} & \textbf{F-Score} \tabularnewline
    \midrule
    Task-Consistency only & 64.1\\
    Cross-Modal Similarity only & 61.0\\
    Both & 67.9\\
    \bottomrule
    \end{tabular}
\label{tab:pseudo}}
\end{subtable}
\hspace{0.0125\linewidth}
\begin{subtable}{0.48\textwidth}{
\centering
\caption{\small{\textbf{Using Step-Localization Annotations.} We compare pseudo summaries from step-localization annotations with our approach.}}
\begin{tabular}{lccc}\toprule
    Method & F-Score \\
    IV-Sum (Step Localization) & 57.6 \\
    IV-Sum (Ours) & 66.8\\
    \bottomrule
    \end{tabular}
\label{tab:stepL}}
\end{subtable}
\end{table*}

\noindent\textbf{Ablations.} We compare different approaches to generate pseudo summaries for training our instructional video summarizer network -- (i) First, we ablate the two objectives, Task Relevance and Cross-Modal Saliency, used to generate the pseudo summaries. (ii) Next, we replace the annotations from our pseudo summary generation pipeline with step localization annotations. We include model and loss ablations in the supplemental.

\noindent\emph{(i) Ablating Objectives.} We ablate the two objectives, Task Relevance and Cross-Modal Saliency, used for generating pseudo summaries, in Tab.~\ref{tab:pseudo}. We train IV-Sum on different versions of pseudo summaries and report F-Scores on the \emph{WikiHow Summaries} validation set. Combining both objectives is more effective than using each objective individually.

\noindent\emph{(ii) Using Step Localization Annotations.} COIN and CrossTask datasets contain temporal localization annotations of a generic set of steps pertaining to the task in the videos. We use these annotations to extract the visual segments corresponding to the steps and concatenate them to form a summary. We assign binary labels to each frame, depending on whether they belong in the summary or not. We then use these step-localization summaries as supervision to train our model, IV-Sum with a weighted-CE loss~\cite{narasimhan2021clip} as this works best for binary labels. In Tab.~\ref{tab:stepL}, we compare this to IV-Sum trained on pseudo summaries generated using our pipeline and report F-Scores on our\emph{WikiHow Summaries} validation set.
As seen, IV-Sum trained on our generated summaries outperforms IV-Sum trained using step-localization summaries. We qualitatively compare our automatic pseudo summaries to the manually labeled step localization annotations in Fig.~\ref{fig:oursvsstepL}. Often the step annotations only cover a few steps and miss other crucial steps as shown in \textcolor{yellow}{yellow} in (1). In (2), we observe that our pseudo summary retrieves steps that are unique to the task which the step localization annotation doesn't include.

    

\section{Conclusion}

We introduce a novel approach for generating visual summaries of instructional videos --- a practical task with broad applications. Specifically, we overcome the need to manually label data in two important ways. For training, we propose a weakly-supervised method to create pseudo summaries for a large number of instructional videos. For evaluation, we leverage WikiHow (its videos and step illustrations) to automatically build a \emph{WikiHow Summaries} dataset. We manually verify that the obtained summaries are of high quality. We also propose an effective model to tackle instructional video summarization, IV-Sum, that uses temporal 3D CNN representations, unlike most prior work that relies on frame-level representations. We demonstrate that all components of the proposed approach are effective in a comprehensive ablation study. 

\noindent\textbf{Acknowledgements:} We thank Daniel Fried and Bryan Seybold for valuable discussions and feedback on the draft. This work was supported in part by DoD including DARPA's LwLL, PTG and/or SemaFor programs, as well as BAIR's industrial alliance programs.

\clearpage

\newpage
\clearpage
\begin{center}
{\bf {\Large Supplementary Materials}\\
TL;DW? Summarizing Instructional Videos with Task Relevance \& Cross-Modal Saliency}
\end{center}

This section is organised as follows:
\begin{enumerate}
    \item \emph{WikiHow Summaries} Data Collection 
    \item Implementation Details
    \item Additional Results 
    \begin{enumerate}
        \item Results on instructional videos in generic video summarization datasets
        \item Step recall 
        \item Model architecture ablations
    \end{enumerate}
    \item Additional Qualitative Results 
    \begin{enumerate}
        \item Qualitative comparison of ground-truth, IV-Sum, CLIP-It, and Step AV
        \item Pseudo summary vs IV-Sum summary
        \item Pseudo summary vs step-localization annotations
        \item Failure case
    \end{enumerate}
\end{enumerate}

\begin{figure*}[h]
    \centering
    \includegraphics[width=\textwidth]{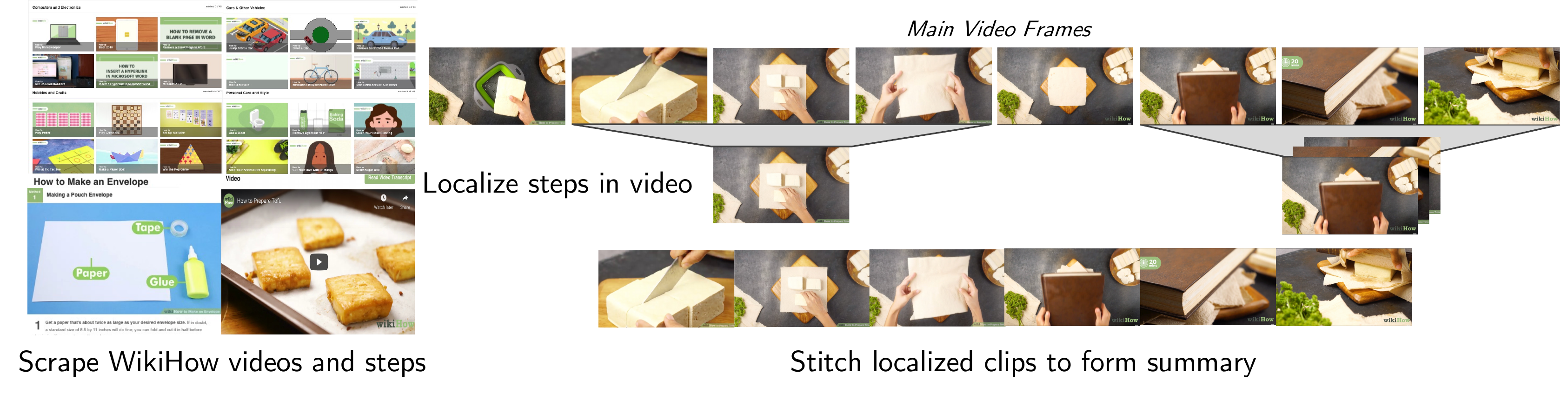}
    \caption{\textbf{\emph{WikiHow Summaries} Data Collection.} We first scrape all the main videos in the WikiHow articles, along with the images or video clips assosciated with each step. Next, the image/clip corresponding to each step is localized in the main video. The images are localized to $\pm$ 2.5 seconds(i.e. a 5 seconds window centered around the image). The localized clips are stitched together to form the summary.}
    \label{fig:dataset}
\end{figure*}
\section{\emph{WikiHow Summaries} Data Collection}

We provide more details on the WikiHow Summaries data collection process. As described in Sec.~4.2 of the main paper, these are the main stages of the dataset creation: (1) Scraping WikiHow videos (2) Localizing images/clips in video (3) Ground-truth summary from localized clips (4) Manual verification. Fig.~\ref{fig:dataset} illustrates our data collection process. We show an example for the article \href{https://www.wikihow.com/Prepare-Tofu}{\emph{``Prepare Tofu''}}. We localize each of the individual steps (images/clip) in the main video by comparing the ResNet features and obtain short localized clips. The clips are stitched together to form the summary. A handful of summaries with spurious lengths (too long or too short) are manually verified and corrected.

We describe how we handle some edge cases in the articles, and the reasoning behind using ResNet features in stage (2).  

\noindent\textbf{Multiple methods.} Sometimes the articles contain multiple methods of performing a task. If the video also contains multiple methods, as in this \href{https://www.wikihow.com/Draw-a-Cow}{\emph{``Draw a cow''}} example, we localize each method in the video, and the summary is a compilation of all methods. The reasoning behind doing this is that users looking for a specific way of drawing a cow can take a quick glimpse of the summary and decide if they want to watch the whole video. However, if the article contains multiple methods but the video only contains one, as in \href{https://www.wikihow.com/Make-an-Envelope}{\emph{this}} example, only the method depicted in the video is added in the summary. 

\noindent\textbf{Reason for using ResNet features instead of direct pixel comparison.} As described in the main paper Sec~4.2, we compare ResNet features to localize the images/clips of the steps in the main video. The reason we compare ResNet features and not pixel values directly is because the images/clips associated with the steps aren't always extracted from the main video. For example, in this article on  \href{https://www.wikihow.com/Make-a-Pinwheel}{\emph{``Making a pinwheel''}}, the frames in the images/clips are from a different video and don't have exact matches in the \href{https://youtu.be/J4_Wuq_VmOY}{\emph{main video}} for the article. Using ResNet features in place of pixels makes the localization robust to color/background changes, allowing us to localize steps despite an exact match of frames.  

\begin{table}[h]
    \centering
    \caption{Hyperparameters for training IV-Sum.}
    \begin{tabular}{l c c}
        \toprule
        \textbf{Hyperparamter} & \textbf{Value} \\
        \midrule
        Batch size & 24\\
        Epochs & 300 \\
        Learning rate IV-Sum & 1e-3 \\
        Learning rate S3D fine-tuning & 1e-4\\
        Weight decay & 1e-4\\
        Dropout & 0.1\\
        Learning rate decay & StepLR\\
        $t\%$ & 55\% & \\
        \#frames per segment & 32\\
        \#frames per video during training & 768\\
        \# Training FPS & 8 \\
        \bottomrule
    \end{tabular}
    \label{tab:hyper}
\end{table}

\section{Implementation Details}
\noindent\textbf{Video processing.} For generating pseudo summaries and for training IV-Sum, the videos are down-sampled to 8 FPS, and divided into non-overlapping segments of size 32 frames, which is the recommended segment size for MIL-NCE~\cite{miech2020end}\footnote{We use the implementation of MIL-NCE available here \url{https://github.com/antoine77340/MIL-NCE_HowTo100M}}. While training IV-Sum, we fix the number of segments sampled from a video to be 28 (i.e. 896 frames) which are selected as a contiguous sequence from a randomly chosen start location. If the video is shorter in duration, it is padded with zeros. During inference, we retain the original fps of the video and all the segments are passed to IV-Sum. For concatenating the text representations to the visual representations, we follow the approach in MIL-NCE and map each visual clip to the sentences a few seconds before, after, and during the clip. The text embedding is an average of all the sentence embeddings. 

\noindent\textbf{Hyperparameters.} Tab.~\ref{tab:hyper} shows detailed list of hyperparameters. For all baselines and our method, to ensure a fair comparison we generate the summary from scores by selecting the top $t\%$ of the highest scoring segments to be in the summary. $t$ is set to be $55$ based on the statistics in the validation set of WikiHow Summaries, where on  average $55\%$ of the original video appears in the summary. 

\noindent\textbf{Dimensions.} We first describe the dimensions of each of the embeddings. The image embeddings are in $f_\text{vid}(s_i) \in \mathbb{R}^{512}$. The text embeddings for $M$ transcript sentences using $f_\text{text}$ are in $\mathbb{R}^{M \times 512}$ which are then fused using a 2 layer perceptron to $\mathbb{R}^{512}$. $M$ is set to be the maximum number of sentences found in any ASR transcript. The image and text embeddings are concatenated and passed to the segment scoring transformer $f_{\text{trans}}$, the output dimension of this is in $\mathbb{R}^{512}$.  

\noindent\textbf{Computation resources.} The training time is approximately 2 days using Distributed Data Parallel to train for 300 epochs on 8 NVIDIA RTX 2080 GPUs. The model inference time for a single video at its original fps is 1.5 minutes on average.

\section{Additional Results}
\begin{table}[h]
    \centering
    \caption{\textbf{Evaluating on generic video summarization datasets.} We compare F-Score of IV-Sum and CLIP-It on the instructional videos in TVSum.}
    \begin{tabular}{lcc}\toprule
        \textbf{Method} & \textbf{F-Score} \\\midrule
        CLIP-It~\cite{narasimhan2021clip} & 0.72\\
        IV-Sum & 0.73\\\bottomrule
    \end{tabular}
    \label{tab:generic}
\end{table}

\noindent\textbf{Evaluating on instructional videos in generic video summarization datasets.} 
Here, we consider the existing generic video summarization datasets, in particular, those videos that fall under ``instructional'' domain, in order to validate our model further.
Generic video summarization dataset TVSum~\cite{Song15} has 15 videos pertaining to the categories \emph{changing a car tire}, \emph{getting a car unstuck}, and \emph{making a sandwich} while SumMe~\cite{Gygli14} has no instructional videos. We follow the evaluation protocol described in CLIP-It~\cite{narasimhan2021clip}. For a fair comparison to CLIP-It~\cite{narasimhan2021clip}, we curate a test set by randomly selecting 7 of these 15 videos, while the remaining 8 are added to the training set, so as to ensure that the CLIP-It model sees instructional videos during training. The augmented training set is curated by combining the 8 videos with those in SumMe (25 videos), TVSum (45 videos), OVP~\cite{OVP} (50 videos), and YouTube~\cite{de11} (39 videos). CLIP-It is trained on this augmented training set consisting of 168 videos (including 8 instructional videos) and evaluated on the held out 7 instructional videos. Our method is trained on pseudo summaries (built on top of CrossTask and COIN) and evaluated in a \emph{zero-shot} way on the test set of 7 videos.

As seen in Tab.~\ref{tab:generic}, our method IV-Sum, although trained with noisy / weakly labeled pseudo summaries from a different data distribution, achieves an F-Score comparable to the CLIP-It~\cite{narasimhan2021clip}, trained on human annotated summaries. 

\begin{table}[h]
    \centering
    \caption{\textbf{Comparing step-recall.} We report step-recall on our method and 2 baselines.}
    \begin{tabular}{lcc}\toprule
        \textbf{Method} & \textbf{Step-recall} \\\midrule
        Step Cross-Modal Similarity & 0.68\\
        CLIP-It with ASR & 0.70\\
        IV-Sum & 0.94\\\bottomrule
    \end{tabular}
    \label{tab:step-recall}
\end{table}

\noindent\textbf{Step recall.} We define an additional metric, \emph{step-recall} to be the average percentage of steps present in the ground-truth summary which were successfully picked by the generated summary. Our \emph{WikHow Summaries} dataset contains annotations of frames pertaining to each step, and if any of the frames from a step are present in the summary, we assume the step is covered. Using this logic, we generate a list of steps in the generated summary $Y'_\text{step}$, and a list of steps in the ground-truth $Y_\text{step}$. We compute step-recall as follows,    

$$\text{Step-recall} = \frac{\text{overlap between } Y_\text{step} \text{ and } Y'_\text{step}}{\text{total duration of } Y_\text{step}}$$

In Tab.~\ref{tab:step-recall} we report the step-recall for Step Cross-Modal Similarity, CLIP-It with ASR (trained on generic video summarization datasets) and IV-Sum. Both Step Cross-Modal Similarity and CLIP-It with ASR baselines miss 30\% of steps found in the ground truth summary while our method on average only misses 6\% of the steps.

\begin{table*}[t]
\centering
\caption{\small{\textbf{Instructional Video Summarizer Ablations.} We perform ablations on different components of the video summarizer network and report results on the \emph{WikiHowTo Summaries} validation set.}}
\begin{subtable}{0.48\textwidth}{
\centering
\caption{\small{\textbf{IVSum S3D backbone Ablations.} We compare fixing the pre-trained weights of the S3D model to fine-tuning a part of it.}}
\begin{tabular}{lcc}\toprule
    \textbf{Method} & \textbf{F-Score} \tabularnewline
    \midrule
    S3D fixed & 65.8 \\
    S3D fine-tuned & 67.9\\
    \bottomrule
    \end{tabular}
\label{tab:s3d}}
\end{subtable}
\hspace{0.0125\linewidth}
\begin{subtable}{0.48\textwidth}{
\centering
\caption{\small{\textbf{IVSum Segment Scoring Transformer Ablations.} We compare different architecture configurations of the segment scoring transformer.}}
\begin{tabular}{lcccc}\toprule
    \multicolumn{3}{c}{\textbf{Method}} & \textbf{F-Score} \tabularnewline
    \midrule
    & \#heads & \#layers & \\
    SST & 8 & 16 & 63.1\\
    SST & 16 & 6 & 63.5\\
    SST & 8 & 12 & 66.7\\
    SST & 8 & 24 & 67.9\\
    MLP & - & - & 32.1\\
    \bottomrule
    \end{tabular}
\label{tab:trans}}
\end{subtable}
\end{table*}

\begin{table*}[t]
\centering
\caption{\small{\textbf{Loss Ablations.} We ablate different losses in the IV-Sum model and show results on validation set of \emph{WikiHow Summaries}}}
\begin{tabular}{lccc}\toprule
    \textbf{Method} & \textbf{F-Score} & \textbf{Recall} \tabularnewline
    \midrule
    MSE & 67.9 & 84.5 \\
    MSE + Diversity & 61.2 & 63.4 \\
    MSE + Reconstruction & 67.6. & 85.8\\
    \bottomrule
    \end{tabular}
\label{tab:loss}
\end{table*}

\noindent\textbf{Loss Ablations.} In Table~\ref{tab:loss}, we explore additional loss functions as in prior video summarization works~\cite{Rochan18,Rochan19,park2020sumgraph,narasimhan2021clip}. Diversity loss ensures diversity among the summary segments and the reconstruction loss enforces similarity in representations of the reconstructed summary and the input video. Adding diversity reduced the recall and we notice no improvement on adding reconstruction loss. We believe this may be because frames corresponding to different steps are not always diverse but are still important for the summary.

\noindent\textbf{Model ablations.} Table~\ref{tab:s3d} shows the performance comparisons between freezing the video and text encoding backbone (S3D) vs. fine-tuning part of the network. In Table~\ref{tab:trans}, we ablate the segment scoring transformer (SST) in our model and change the number of encoder layers, heads, and also replace the transformer with an MLP. We report the F-Score on the validation set of \emph{WikiHowTo Summaries}.  

For a fair comparison of MIL-NCE vs CLIP features, we retrain our IV-Sum model replacing video segments with frames and replacing MIL-NCE features with CLIP image and text features (same as the ones used in the CLIP-It baseline). We report results in Tab.~\ref{tab:1}.  We see that IV-Sum with CLIP performs at par with CLIP-It with ASR but falls short of IV-Sum, indicating the need to use video segments and MIL-NCE features pre-trained on HowTo100M. 

\begin{table}[H]
    \centering
    \small
    \caption{\small{\textbf{Instructional Video Summarization results on \emph{WikiHow Summaries.}} All models were trained on pseudo summaries.}}
    \begin{tabular}{lcccc}\toprule
   {\multirow{2}{*}{Method}} & \multicolumn{2}{c}{F-Score} & $\tau$ (Kendall) & $\rho$ (Spearman) \\ 
    \cmidrule(l{1pt}r{2pt}){2-3} 
    & Val & Test & Test & Test\\ \midrule
    CLIP-It with ASR & 62.5 & 61.8 & 0.093 & 0.191 \\ 
    IV-Sum with CLIP & 61.8 & 62.0 & 0.094 & 0.201 \\
    \textbf{IV-Sum} & \textbf{67.9} & \textbf{67.3} & \textbf{0.101} & \textbf{0.212}\\
    \bottomrule
    \end{tabular}  
\label{tab:1}
\end{table}

\section{Additional Quantitative Results}

Please watch the video on our \href{https://medhini.github.io/ivsum/}{website} for qualitative results. 

\noindent\textbf{Comparison to baselines.} We show video results comparing the ground-truth summary to that from IV-Sum (our method) and baselines Step Cross-Modal Similarity and CLIP-It with ASR trained on generic video summarization datasets. Our method picks all frames in the ground-truth and assigns high scores to salient frames. Step Cross-Modal Similarity misses the crucial step \emph{``fold and tuck''} at the end as it assigns higher scores to irrelevant frames at the start of the video. This is because it has no knowledge of task-relevance. CLIP-It with ASR (trained on generic video summarization datasets) misses steps (like \emph{``fold into a triangle''}) and assigns lower scores to the key frames in a step as it optimizes for diversity. 
\noindent\textbf{Evaluating Pseudo Summary generation procedure for WikiHow Summaries.} We found that there are 15 tasks which are shared between the Pseudo Summary training set and the WikiHow Summaries test set. We applied the method used to construct pseudo summaries to these 15 task videos in the WikiHow Summaries by fetching videos of the same task from our training set. We compare this to IV-Sum and report results in ~Tab \ref{tab:2}. We notice a slight improvement on all three metrics, indicating that our model is able to learn above the noise in the pseudo summaries. 

\begin{table}[H]
    \centering
    \small
    \caption{\small{\textbf{Evaluating pseudo summary generation on subset of WikiHow Summaries} }}
    \begin{tabular}{lccc}\toprule
   {\multirow{2}{*}{Method}} & F-Score & $\tau$ & $\rho$ \\ 
    \cmidrule(l{1pt}r{2pt}){2-3} 
    & Test & Test & Test\\ \midrule
    Pseudo Summary Generation & 38.0 & 0.03 & 0.36  \\ 
    \textbf{IV-Sum} & \textbf{42.0} & \textbf{0.04} & \textbf{0.38} \\
    \bottomrule
    \end{tabular}  
\label{tab:2}
\end{table}

\noindent\textbf{Pseudo summary vs IV-Sum summary.} IV-Sum is trained on weakly labeled pseudo summaries that may sometimes be noisy. However, since the training loss is not 0, we check if our model learns despite the noise and produces summaries of a better quality. In this example, we show summaries for \href{https://www.youtube.com/watch?v=9krpJsOi3dE}{\emph{``this''}} video from the Pseudo Summaries dataset. As seen the pseudo summary contains an irrelevant segment where results from a web search are shown in Korean for nearly 10 seconds ($9^{\text{th}}$ second to the $19^{\text{th}}$ second). The IV-Sum model trained on pseudo summaries yields a resulting summary without this segment, as it is able to  learn \emph{``task-relevance''} and \emph{``cross-modal saliency''}. 



\looseness=-1
\noindent\textbf{Pseudo summary vs step-localization annotation.} We compare pseudo summaries generated using our method to the step-localization summary. In the example \emph{``Make a pumpkin spice latte''}, the input video can be found \href{https://www.youtube.com/watch?v=9krpJsOi3dE}{\emph{here}}. Step localization only localizes two main steps, \emph{``boil milk''} and \emph{``add coffee and blend''} whereas our pseudo summary contains all the main steps necessary to do the task.

\noindent\textbf{Failure case.}  
Since we always select the top $55\%$ of the segments to be in the summary (i.e. t=55\%), the summary chosen by our method is sometimes much longer/shorter than the ground-truth summary. This is a failure case of the baseline methods as well. For example, for this 38 second video on \href{https://www.youtube.com/watch?v=HX1okLOSRz8}{\emph{``How to ripen a cantaloupe''}}, the ground-truth summary is a brief 15 seconds whereas our summary covers the steps in more detail and is 21 seconds long.

\clearpage
%
%

\bibliographystyle{splncs04}
\bibliography{egbib}

\end{document}